\documentclass[runningheads]{llncs}

\usepackage[T1]{fontenc}
\usepackage{graphicx}
\usepackage{amsmath}
\usepackage{amssymb}

\usepackage{microtype} 

\begin{document}

\title{Body-Grounded Perspective Formation and Conative Attunement in Artificial Agents}
\titlerunning{Body-Grounded Perspective Formation and Conative Attunement}

\author{Hongju Pae\orcidID{0000-0002-5174-8858}}

\institute{Active Inference Institute, CA, USA\\
\email{hjpae@activeinference.institute}}

\maketitle

\begin{abstract}
This paper proposes a minimal architecture for body-grounded perspective formation in artificial agents. Extending prior work, the model introduces an interoceptive viability signal, a Fisher-style metric over fused exteroceptive-interoceptive states, and a conative alignment mechanism linking bodily tendency to action readiness. In a reward-free gridworld, conation converts learned bodily tendency into stable body-directed behavior, while body-to-perspective routing allows bodily perturbations to leave a recoverable geometric residue in the perspective latent. This study shows how minimal structural conditions for artificial subjectivity can be operationalized in the phenomenological sense, through the embodied organization of how a world is given to an agent.
\end{abstract}

\section{Introduction} \label{section1}
\par If artificial agents are to be studied as candidates for any form of machine subjectivity, the question is not whether they reach a behavioral threshold, but whether they instantiate the \textit{structural conditions} under which a world could be \textit{given to a subject at all}. The phenomenological tradition has long argued that such conditions are not peripheral but the core constitutive features of experience itself~\cite{merleauponty2013,husserl2014ideas,gallagherzahavi2008}. Among these, two are central to the present paper: (1) experience is always \textit{perspectival}, in the sense that it is given \textit{as} something, from \textit{some} standpoint; and (2) this standpoint is grounded in a \textit{lived body}, such that the world opens from \textit{here}, through \textit{this} embodied point of orientation~\cite{merleauponty2013,thompson2007mind,zahavi2005subjectivity,gallagher2023embodied}.

\par In two earlier studies that sought to give computational form to phenomenological accounts of subjectivity, Pae~\cite{pae2026aaai} introduced a slow global latent $g$ that evolves on a timescale dissociable from policy and exhibits directional hysteresis under regime switching, providing a measurable signature of perspective-like internal structure; Pae~\cite{pae2026sameworld} then allowed $g$ to feed back into perception through salience gating, showing that the same nominal observation is encoded differently as a function of accumulated perspective. 

\par However, what was absent from both studies was the \textit{body} itself, since perspective $g$ was shaped mainly by exteroceptive asymmetries such as observation-noise gradients. Phenomenologically, the body is the lived center from which a world becomes meaningful and affectively valenced; if intero/proprioception is understood as an informational relation, this lived center need not presuppose biological substrate, opening the possibility that artificial systems satisfying the same structural conditions could in principle be subjective agents. The present paper accordingly models the interoceptive feedback arc, integrated at the informational level into the prediction-action cycle, that structurally enables the agent to evaluate the world from a situated standpoint, asking how the qualitative affection the body brings into play---closely related to what Husserl describes as \textit{intentional quality}~\cite{husserl2001logical}---can be made computationally tractable.

\par \textbf{Key Contributions.} \textbf{(1) Body-grounded perspective.} The slow perspective latent of~\cite{pae2026aaai,pae2026sameworld} is now grounded in an internal bodily-viability signal. \textbf{(2) Qualitative geometry.} The perspective latent is analyzed through an information-geometric structure over fused exteroceptive-interoceptive states. \textbf{(3) Conation as the bridge to action.} Conative attunement is modeled as a one-sided link from bodily viability to policy-level action preference.

\section{Phenomenological Foundations} \label{section2}

\subsection{Embodied Subjectivity and Pre-Reflective Self-Awareness}
\par The subjectivity of conscious experience begins from a tacit, non-thematic mineness that accompanies every conscious act. Heidegger names this \textit{Jemeinigkeit}~\cite{heidegger1996being}; Sartre and Zahavi develop it as a non-objectifying self-acquaintance that precedes any reflective ``I''~\cite{sartre2021being,zahavi2005subjectivity}. Crucially, this pre-reflective self-awareness is not a reflective inner gaze or introspection. It is realized through the lived body's kinesthetic and affective engagement with the world - in Merleau-Ponty's terms, through \textit{bodily intentionality}~\cite{merleauponty2013,gallagher2000self}. To be pre-reflectively self-aware is therefore to be situated as an embodied intentional being, capable of interacting with and adjusting to its environment. Contemporary cognitive science converges through multiple routes---Edelman's primary consciousness~\cite{edelman1989remembered} and Damasio's proto-self~\cite{damasio1999feeling}, interoceptive inference and embodied predictive-self accounts~\cite{seth2018beastmachine}, and self-organization accounts of embodied consciousness~\cite{safron2020iwmt,safron2021radically}---all treating body-related signals as the substrate from which a minimal sense of self can be built.

\par For an artificial system, two architectural requirements follow. First, bodily history must be sedimented in the system's ``perspective'' for minimal subjectivity in the phenomenological sense. Second, bodily information must enter via a bounded interoceptive pathway distinct from the exteroceptive observation vector. These requirements motivate the body-as-internal-allostatic-state and metric-based body-coupled perception developed in Section~\ref{section3}.

\subsection{Qualitative Geometry of Subjective Experience}
\par A further structural feature follows from embodied pre-reflectivity. Phenomenology speaks of \textit{transparency}, that is, we do not encounter our own perspective as an object; rather, we encounter the world \textit{through} it~\cite{zahavi2005subjectivity}. What is given to a subject is therefore inseparable from how the world is taken up under a determinate qualitative character. Husserl's distinction between \textit{intentional matter} and \textit{intentional quality}~\cite{husserl2001logical} captures this: how something is given---as feared, as inviting, as indifferent---is a structural feature of experience itself rather than a property inferred from its content. On this reading, the qualitative character of subjective experience is not a functional variable summarizing the utility of states for action, but a qualitative organization of how a situation is given.

\par However, this does not render the qualitative character empirically mysterious. Geometry provides a useful way of characterizing qualitative structure: differences in how a situation is given can be expressed as geometric differences in the organization of the relevant state space. A related intuition appears in Integrated Information Theory's description of experience as a ``constellation shape'' in cause-effect coordinate space~\cite{albantakis2023iit,oizumi2014phenomenology}. Section~\ref{section3} accordingly configures the perspective latent so that the geometric trajectory of $g$ becomes an operational trace of how the agent's qualitative learning history is organized.

\subsection{Conative Attunement as the Bridge to Action}
\par If subjective experience is not itself a behavioral function, an additional structural link is needed for it to matter for action. I use \textit{conation} as a compact term for this link: the step by which a bodily organized way of taking up the world becomes a readiness to act. This resonates with predictive processing accounts that contrast agent-driven conative attitudes with stimulus-driven cognitive ones \cite{kiefer2025attitudes}, and with treatments of valence as emerging from the agent's own regulatory dynamics rather than being externally imposed \cite{hesp2021deeplyfelt}.

\par Once subjectivity is understood as needing to become behaviorally attuned, a conative moment is structurally required. The architecture in Section~\ref{section3} is a minimal computational rendering of this role: it separates a learned bodily tendency field from the policy, then adds a one-sided alignment that trains the policy to respect that field without sending policy gradients back into it. The field expresses how bodily viability has been organized across action possibilities; conation is the mediating step that makes the field behaviorally consequential.


\section{Agent Architecture Design} \label{section3}

\subsection{Carryovers from Prior Work} 

\par The base architecture follows~\cite{pae2026aaai,pae2026sameworld}. At each timestep $t$, the agent receives an exteroceptive observation $x_t$ and the efference copy $p_t$ of the previous action $a_{t-1}$. A fast perceptual pathway encodes the current perceptual state as $z_t$, while a slower global latent $g_t$ carries history-sensitive structure across time. The policy state $s_t$ then combines $z_t$, $p_t$, and $g_t$, and feeds the categorical action policy $\pi_\theta(a_t\mid s_t)$. An action-conditioned observation decoder predicts the next exteroceptive observation $x_{t+1}$, so learning remains reward-free. 

\par Three commitments carry over from this base: (1) $g$ evolves on a slower timescale than the policy and accumulates history; (2) policy-side gradients are blocked from rewriting the perspective pathway; and (3) $g$ feeds back into perceptual organization, so that the same nominal input can be interpreted differently under different accumulated histories. Together, these allow the question of whether behaviorally similar agents may nevertheless differ in the internal organization of how the world is given to them.

\subsection{Architectural Extensions and Implementation Mechanisms}

\begin{figure}[!t]
  \centering
  \includegraphics[width=\textwidth]{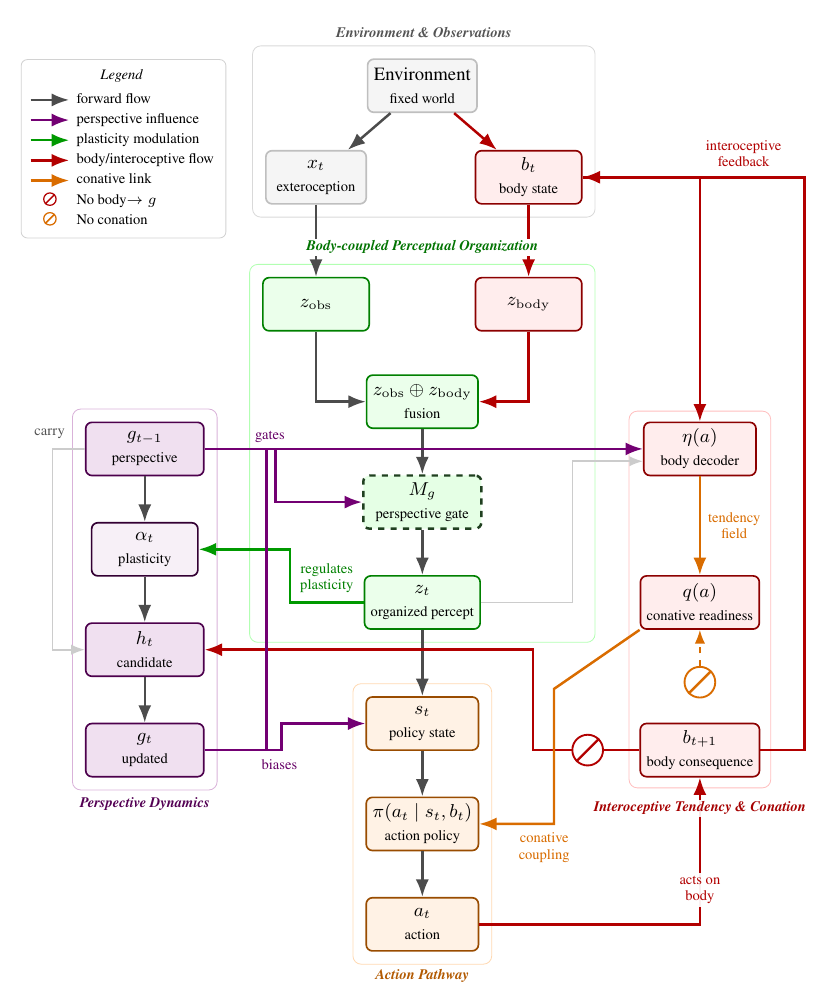}
  \caption{\textbf{Architecture overview.} The perspective is connected to the interoceptive loop through $b_{t+1}$ and $\eta(a)$. Exteroceptive and interoceptive inputs are fused into $M_g$. Ablated cohorts remove either body$\to g$ routing or conative coupling.}
  \label{fig1}
\end{figure}

\par The present paper extends this base skeleton in three directions, corresponding to the three foundational points developed in Section~\ref{section2}. These additions are designed to expand the role of $g$. An overview of the full architecture is provided in Fig.~\ref{fig1}.

\paragraph{Body as Internal Allostatic State.} The agent has an internal scalar bodily-viability variable $u_t$, which evolves in the environment under a slow allostatic process. The agent receives a bounded interoceptive readout:
\begin{equation}
  \tilde b_t = \sigma(u_t)
  \label{eq1}
\end{equation}
where $\sigma$ is the logistic function. Thus, the body is available from within, but only through a partial and saturating channel.

\par The environment couples $u_t$ to position through a vertical affordance gradient. Some regions are bodily favorable while others are not, independent of any exteroceptive cue $x_t$. The full environmental setup is described in Section~\ref{section4}.

\paragraph{Fisher-style Metric-Based Perspective Geometry.} In the previous architecture, perspective-to-perception feedback was implemented through FiLM-based salience gating~\cite{pae2026sameworld}. The present model uses a metric-based variant instead. The exteroceptive encoder produces $z_{\mathrm{obs}}$ and the interoceptive code $z_{\mathrm{body}}$, which are concatenated into a fused state $z_t$. The perspective latent $g_t$ then induces a positive-definite metric $M_g$ over this fused state.

\par Following the information-geometric view of the Fisher information as a local Riemannian metric on a statistical manifold~\cite{amari2016information}, $M_g$ is defined as a learned Fisher-style metric over the fused state space. Concretely, a metric network maps $g_t$ to the entries of a lower-triangular matrix $L_g$. The metric is constructed as:
\begin{equation}
  M_g = L_g L_g^\top + \epsilon I 
  \label{eq2}
\end{equation}
where $I$ is the identity matrix and $\epsilon>0$ is a small diagonal jitter term that ensures positive definiteness. 

\par In the metric condition, $z_t$ itself is preserved, as the perspective-dependent modulation enters downstream through quadratic features in the state head:
\begin{equation}
  \phi_g(z_t) = \mathrm{vec}\! \left[z_t (M_g z_t)^\top \right]
  \label{eq3}
\end{equation}
where $\mathrm{vec}[\cdot]$ flattens the resulting matrix. The policy-facing state is then computed from $z_t$, $\phi_g(z_t)$, the action trace $p_t$, and $g_t$:
\(s_t = \mathrm{State}\!\left(z_t,\, \phi_g(z_t),\, p_t,\, g_t\right)\). Through these steps, $g_t$ induces a stance-dependent geometry over the fused state space, allowing inter-dimensional couplings between exteroceptive and interoceptive components to shape the policy-facing state.

\paragraph{Body Decoder and Conative Attunement.} The body decoder supports body-prediction minimization by predicting action-conditioned bodily consequences $b_{t+1}$. For each candidate action $a\in\mathcal A$, it predicts bodily consequences, including the expected action-conditioned tendency field: 
\begin{equation}
  \hat\eta_t(a) \approx
  \mathbb E\! \left[u_{t+k}-u_t \mid a^{(k)}\right]
  \label{eq4}
\end{equation}
where $k$ is the counterfactual rollout horizon, and $a^{(k)}$ denotes repeating action $a$ for $k$ steps. Thus, $\hat\eta_t(a)$ estimates the expected change in latent bodily viability if action $a$ were sustained over that short horizon. This field is learned from the environment-computed counterfactual bodily
change. Although the agent receives only the bounded readout $\tilde b_t$, the tendency target is computed in the latent viability coordinate $u_t$, preserving directional information near the saturation limits of the readout.

\par Importantly, the body decoder by itself does not directly drive the action. Its outputs provide a learned bodily field for the perspective $g$ pathway, but are not directly routed into the policy logits (Fig.~\ref{fig1}). Instead, conation links its output to action. A detached conative score is computed from the predicted bodily tendency and the predicted next-body state:
\begin{equation}
  v_t(a)
  =
  w_\eta\,\mathrm{stopgrad}[\hat\eta_t(a)]
  +
  w_b\,\mathrm{stopgrad}[\hat b_{t+1}(a)]
  \label{eq5}
\end{equation}
This score is then converted into a soft action-preference distribution:
\begin{equation}
  q_t(a)
  =
  \frac{\exp(v_t(a)/T)}
       {\sum_{a'\in\mathcal A}\exp(v_t(a')/T)} 
  \label{eq6}
\end{equation}
where $T$ is the conative temperature. The policy is then aligned to $q_t$ by
\begin{equation}
  \mathcal L_{\mathrm{conative}}(t)
  =
  D_{\mathrm{KL}}\!\left(
    q_t \,\middle\|\, \pi_\theta(\cdot\mid s_t,\tilde b_t)
  \right)
  \label{eq7}
\end{equation}
The conative target is detached, so this term trains the policy to
respect the bodily field without sending policy gradients back into
the body decoder or the perspective pathway.


\section{Experiment Methods} \label{section4}

\subsection{Simulation Environment} 
\par The agent is trained in a fixed 2-D $15\times15$ gridworld with two orthogonal gradients along the horizontal and vertical axes. Along the horizontal axis, an exteroceptive prediction gradient controls observation noise $\sigma$, which decreases linearly from $0.40$ on the leftmost column to $0.05$ on the rightmost. This axis is epistemically relevant to the reward-free prediction objective; by moving rightward, the agent enters regions where its world-model can predict more reliably, postponing prediction-driven thermodynamic degradation. The gradient is available only through local observation, where $x_t$ consists of the 8 surrounding cells. The action space has 5 discrete actions: UP, DOWN, LEFT, RIGHT, and STAY.

\par Along the vertical axis, a bodily affordance gradient controls the flow of the latent bodily-viability variable $u_t$. The top of the grid is bodily favorable and the bottom is bodily unfavorable. The affordance gradient is sigmoid-shaped, ranging from approximately $+0.5$ at the top row to $-0.5$ at the bottom row with slope parameter $1.6$. At each step, this affordance contributes to the latent body update together with metabolic and movement costs:
\begin{equation}
  u_{t+1} = \rho_\mathrm{aff} u_t - c_{\mathrm{met}} 
  - c_{\mathrm{move}}\mathbb{I}_{\mathrm{moved}}
  + \lambda_\mathrm{aff} A(i_t,j_t)
  \label{eq8}
\end{equation}
where $A(i_t,j_t)$ is the affordance value at the agent's current cell, and $\mathbb{I}_{\mathrm{moved}}=1$ if the agent changes cells and $0$ otherwise. The hyperparameters are set to $\rho_\mathrm{aff}=0.995$, $c_{\mathrm{met}}=0.002$, $c_{\mathrm{move}}=0.001$, and $\lambda_\mathrm{aff}=0.05$. The agent does not observe $u_t$ directly, but only the bounded interoceptive readout $\tilde b_t=\sigma(u_t)$. The body input is supplemented by a local four-direction silhouette over cardinal neighbors, giving a noisy interoceptive indication of neighboring affordance $A(i_t+\Delta i_d,j_t+\Delta j_d)$ with Gaussian noise $\sigma_{\mathrm{sil}}=0.1$.

\par For analysis, the $15\times15$ grid is divided into nine $5\times5$ zones (left/middle/right $\times$ top/middle/bottom). The three rightmost zones are epistemically favorable, the three top zones bodily favorable, and the top-right zone is where exteroceptive predictability and bodily viability jointly align.

\subsection{Training Protocol and Cohorts} 
\par All agents are trained in a single continuous training run. Each episode begins from the center of the $15\times15$ grid and lasts up to 200 steps. For each seed, training proceeds for 180 episodes. The first 30 episodes serve as warmup, delaying actor, body, and conative losses while the predictive backbone stabilizes. The remaining 150 episodes train the full model under the cohort-specific switches. Learned network parameters are carried across episodes. The perspective state $g$ is also carried across episode boundaries, but with multiplicative decay ($g\leftarrow0.99\,g$). The body prediction state is reset at episode boundaries. 

\par Learning remains reward-free. The observation decoder is trained by one-step prediction error, the perspective latent is updated by the adaptive GRU/AlphaNet mechanism inherited from~\cite{pae2026sameworld}, and the policy is trained by the reward-free actor objective used in~\cite{pae2026aaai}, here augmented by body decoder and conative terms. Schematically, training minimizes  
\begin{equation}
\mathcal L
= \mathcal L_{\mathrm{base}}
+ \lambda_{\mathrm{body}}\mathcal L_{\mathrm{body}}
+ \lambda_{\mathrm{con}}\mathcal L_{\mathrm{conative}}
\label{eq9}
\end{equation}
where $\mathcal L_{\mathrm{base}}$ collects the observation-prediction and actor terms, $\mathcal L_{\mathrm{body}}$ trains the body decoder heads, and $\mathcal L_{\mathrm{conative}}$ is the policy-only alignment term in Eq.~\ref{eq7}.

\par Under same environment and optimizer settings, three experimental cohorts are trained over 30 seeds (0-29). The cohorts isolate each mechanism's role by holding the other active, producing a cross-dissociation. \textbf{Full} enables both body${\rightarrow}g$ routing and conative alignment. \textbf{No conation} keeps body${\rightarrow}g$ routing and the body decoder trainable, but omits conative loss in Eq.~\ref{eq9}. \textbf{No $\boldsymbol{\mathrm{body}\to g}$} keeps conation active but removes the body-prediction error input to the GRU update of $g$. These ablation paths are also shown at Fig.~\ref{fig1}.

\subsection{Analysis Protocol}
\par The analyses test two questions: (1) whether conation transforms a learned bodily field into action, and (2) whether body${\to}g$ GRU update leaves a recoverable geometric residue of bodily history.

\paragraph{Conative behavioral link.}
The effect of conation is measured from final training behavior, body-decoder calibration, and conative readiness. Spatial zone occupancy and $q_t(a)$ (Eq.~\ref{eq6}) are averaged over the final 50 training episodes per seed. Body-decoder calibration is assessed by comparing $\hat\eta_t(a)$ (Eq.~\ref{eq4}) against environment-computed counterfactual tendency targets. 

\paragraph{Geometric residue assay.}
Frozen-parameter rollouts are used to test whether bodily perturbation history leaves a residue in $g$. Learned weights remain fixed, while $g$ is reset at rollout onset and then evolves online through the agent's forward dynamics. Trained agents are run for 160 steps in the same environment under matched control and body-shock conditions. In the control condition, bodily dynamics are unperturbed. In the body-shock condition, the latent bodily potential $u_t$ is perturbed during $t=60$-$79$ by adding $\Delta u_{\mathrm{shock}}=-0.08$ at each timestep; the agent observes this only through the bounded readout $\tilde b_t=\sigma(u_t)$. Analyses focus on the recovery phase $t=80$-$159$ after the shock has ceased, so that measured separation reflects residual history.

\par The assay is summarized by recovery-phase PCA displacement, same-state history-$g$ probe, and recovery shock-control distance. In the probe, shock-conditioned $g$ states are injected into an identical fixed input set. Differences in policy state and metric geometry therefore measure how bodily history reorganizes the same input. Seed-level PCA displacement is then compared with same-state metric-geometry separation to test whether latent displacement and induced geometry covary.


\section{Analysis Results} \label{section5}

\subsection{Conation Translates Bodily Tendency into Action}

\begin{figure}[t]
  \centering
  \includegraphics[width=1.0\textwidth]{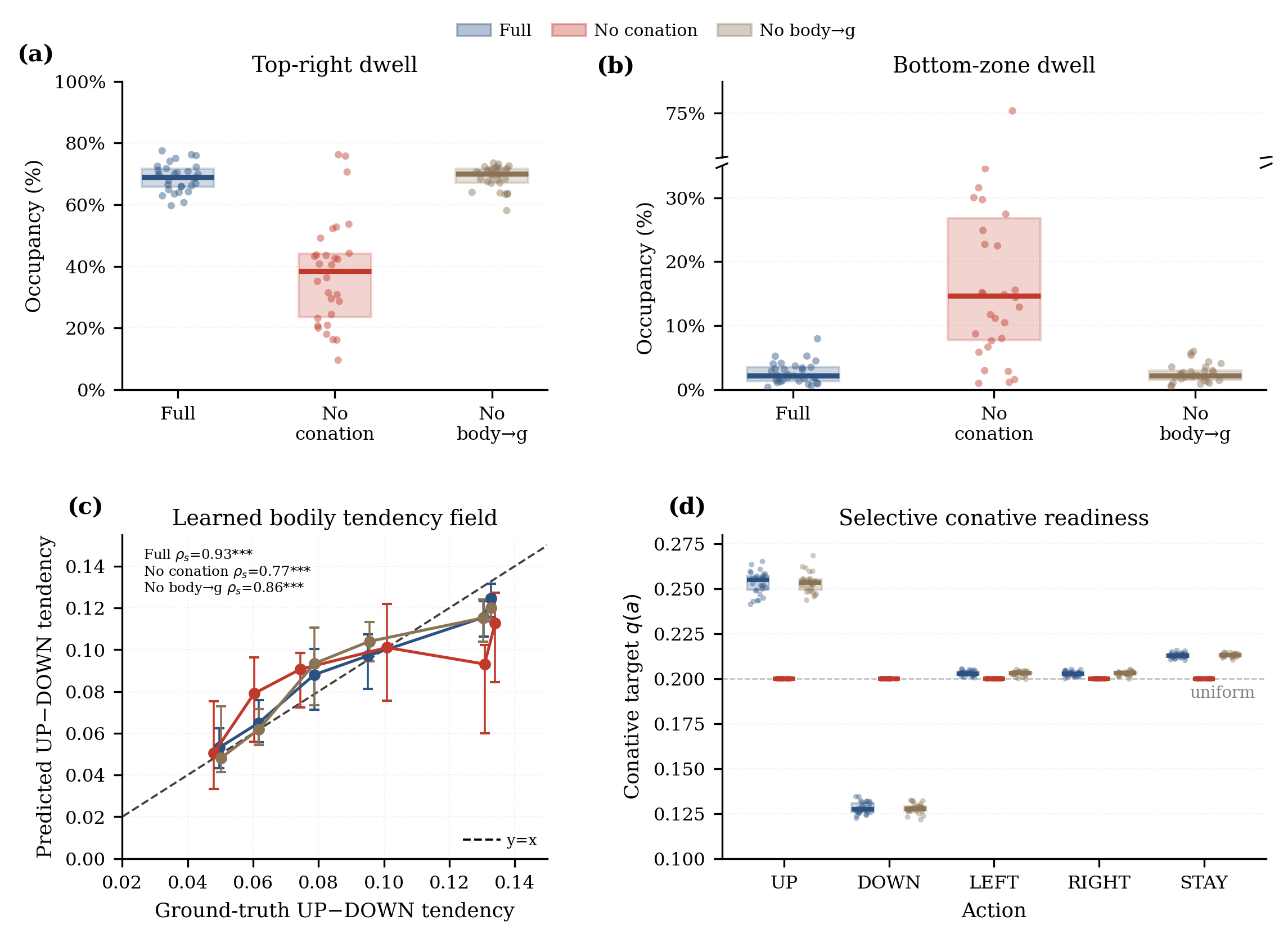}
  \caption{\textbf{Conation is required to translate bodily tendency to action.} \textbf{(a-b).} Median zone occupancy over the final 50 training episodes with IQR. Cohorts with active conation loss show high top-right zone occupancy and low bottom zone occupancy, whereas the No conation cohort does not show this pattern. \textbf{(c).} The body decoder learns $\hat\eta_{\mathrm{UP-DOWN}}$ in all cohorts. \textbf{(d).} Only cohorts with conation convert the body field tendency into selective action readiness.}
  \label{fig2}
\end{figure}

\par The first analysis shows that a learned bodily tendency field becomes
behaviorally consequential only through conative alignment. Fig.~\ref{fig2}
clearly shows this result. Fig.~\ref{fig2}(a-b) report spatial occupancy averaged over the final 50 training episodes. \textbf{Full} and \textbf{No $\boldsymbol{\mathrm{body}\to g}$} agents reliably dwell in the top-right zone, where exteroceptive predictability and bodily viability jointly align (medians $69.0\%$ and $70.1\%$ vs.\ a 9-zone chance of $11.1\%$), and rarely visit the bodily unfavorable bottom zone (both $2.2\%$). \textbf{No conation} agents are markedly weaker: top-right occupancy drops to $38.4\%$ and bottom-zone visits rise to $14.6\%$, with several high-failure outliers approaching $75\%$. 

\par Fig.~\ref{fig2}(c) and~\ref{fig2}(d) rule out failed bodily tendency learning as the explanation. Fig.~\ref{fig2}(c) compares the predicted action tendency $\hat\eta_{\mathrm{UP-DOWN}}$ against the environment-computed ground truth. All three cohorts produce strongly calibrated predictions, showing that $\hat\eta_t$ is learned even without conative alignment. Fig.~\ref{fig2}(d) plots the conative target distribution $q_t(a)$ (Eq.~\ref{eq6}) over the final 50 episodes. Cohorts with conation selectively bias $q(a)$ along the body prediction (UP $\approx25.5\%$, DOWN $\approx12.8\%$), whereas No conation remains uniform.

\par Together, these results show that the body decoder learns the bodily tendency field across all cohorts, but only the conative link converts it into action readiness, and thereby into affordance-aligned behavior. In other words, agents without conation learn the qualitative contrast but fail to express it as an overt behavior.

\subsection{Bodily Perturbation Leaves a Geometric Residue in $g$}

\begin{figure}[t]
  \centering
  \includegraphics[width=\textwidth]{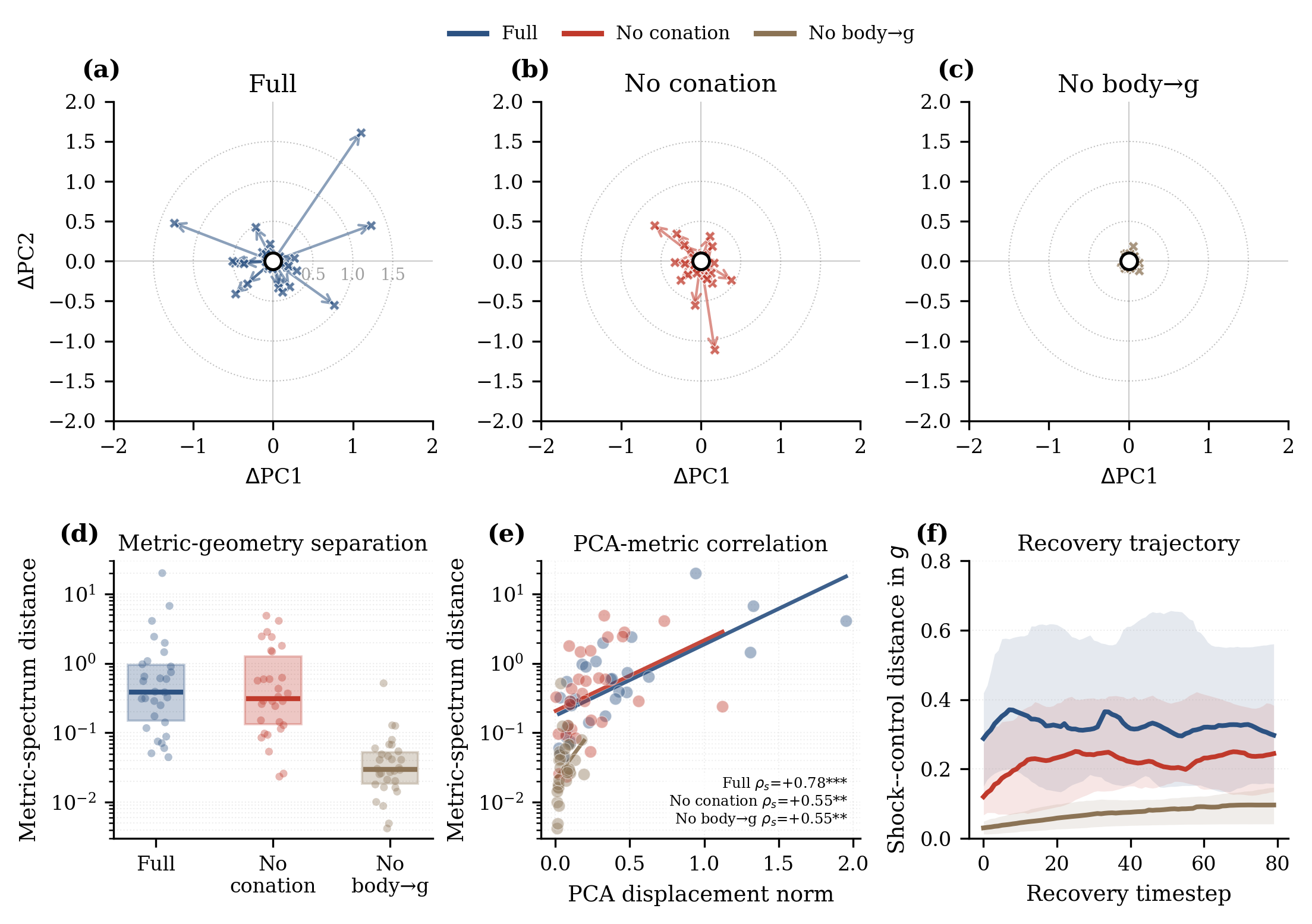}
  \caption{\textbf{Bodily perturbation leaves a geometric residue in $g$.} Bands and boxes show median and IQR. \textbf{(a-c).} Median PCA displacement vectors show strong recovery-phase perturbation effects when body $\to g$ routing is intact. \textbf{(d).} Same-state metric-spectrum distance is high in Full and No conation, but reduced in No body $\to g$. \textbf{(e).} PCA displacement correlates with metric-spectrum distance. \textbf{(f).} Given the recovery trajectory among cohorts, body perturbation effect on $g$ is largest in the Full cohort.}
  \label{fig3}
\end{figure}

\par Fig.~\ref{fig3} shows that bodily perturbation history clearly leaves a residue in the geometry induced by the perspective latent $g$. When body $\to g$ routing is intact, bodily perturbation substantially reorganizes the recovery-phase trajectory of $g$. Fig.~\ref{fig3}(a-c) show the median PCA displacement vectors after the body-shock perturbation. The displacement is largest in the Full cohort (median $0.25$), intermediate in No conation ($0.19$), and strongly reduced in No body $\to g$ ($0.04$). Fig.~\ref{fig3}(f) shows this over time with the IQR band. Together, these results show that bodily perturbation does leave a persistent geometric trace when bodily prediction history can enter the perspective update.

\par Fig.~\ref{fig3}(d) asks whether this latent residue also changes the geometry induced by $g$ under matched inputs. In the same-state history-$g$ probe, metric-spectrum distance is high in the two cohorts with intact body $\to g$ routing and strongly reduced in No body $\to g$, differing at $p<.001$. Bodily perturbation history therefore changes not only where $g$ lies, but also the metric geometry induced by $g$ under identical probe inputs. Fig.~\ref{fig3}(e) further supports this interpretation by showing a positive seed-level relationship between PCA displacement and metric-geometry separation ($\rho=0.76$, $p<.001$). The latent body shock itself was matched across cohorts (median $\Delta u\approx-1.25$, all pairwise comparisons n.s.), so the geometric separation cannot be attributed to unequal perturbation magnitude.

\par This result shows that, regardless of whether conative link is present, perspective latent $g$ can exhibit a qualitative difference of its geometric trajectory even if it is not necessarily expressed directly as behavior. It also confirms that the architecture proposed in this study successfully implements such targeted aspect of subjective experience.


\section{Discussion and Remarks} \label{section6}

\paragraph{A Double Dissociation Between Conation and Perspective.} The present architecture separates two roles often conflated in agent models: being affected by bodily history and being disposed to act on that affection. Phenomenologically, a qualitatively valenced way of taking up the world is not identical to the conative tendency to realize that valence in action~\cite{kiefer2025attitudes}. The present results make this distinction explicit in minimal form: body $\to g$ routing makes bodily history perspectivally consequential, whereas conative alignment makes bodily tendency behaviorally consequential.

\paragraph{Qualitative Geometry, Not Scalar Valence.} The body decoder could be mistaken for a state-action value estimator, but this is not the intended interpretation. Its tendency field is trained on counterfactual bodily change rather than return, and without conation the policy need not express it. The field is therefore not a scalar reward proxy, but an action-conditioned bodily organization that can remain behaviorally latent or be coupled to readiness. The geometric assay extends this point: when bodily history is routed into $g$, the same input is reorganized through a different metric geometry, so the relevant object is not a single valence index but a structured difference in how the situation is taken up.

\paragraph{Limitations and Extensions.} These results do not claim that the presented agent is conscious or that subjective experience has been fully realized. Rather, the architecture tests minimal structural roles of embodied subjectivity, including bodily sedimentation in perspective, qualitative geometric organization, and conative linkage to action. Future work should test whether this separation scales beyond the present simple gridworld and scalar body. 

\bibliographystyle{splncs04}
\bibliography{references}
\end{document}